\documentclass[two-column]{ieeeconf}
\usepackage{graphicx}
\usepackage{amsmath,amssymb,amsfonts,bm}
\usepackage{mathptmx}
\usepackage{subcaption}
\usepackage{booktabs}
\usepackage{cite}
\usepackage{url}
\usepackage{multirow}
\usepackage{float}
\usepackage{color}
\usepackage{mathrsfs}

\newtheorem{Assumption}{Assumption}

\usepackage{gensymb}
\usepackage[ruled,vlined,linesnumbered]{algorithm2e}
\usepackage{indentfirst}

\usepackage{bm}
\title{Multi-Agent Formation Navigation Using Diffusion-Based Trajectory Generation }
\author{
Hieu Do Quang,
Chien Truong-Quoc$^*$\thanks{$^*$Corresponding authors},
and Quoc Tran Van$^*$\thanks{
Department of Mechatronics, School of Mechanical Engineering,
Hanoi University of Science and Technology, Hanoi, Vietnam.
Emails: hieu.dq226325@sis.hust.edu.vn,
chien.truongquoc@hust.edu.vn,
quoc.tranvan@hust.edu.vn
}
}

\begin{document}

\maketitle

\begin{abstract}
This paper introduces a diffusion-based planner for leader--follower formation control in cluttered environments. 
The diffusion policy is used to generate the trajectory of the midpoint of two leaders as a rigid bar in the plane, thereby defining their desired motion paths in a planar formation. While the followers track the leaders and form desired foramtion geometry using a distance-constrained formation controller based only on the relative positions in followers' local coordinates. 
The proposed approach produces smooth motions and low tracking errors, with most failures occurring in narrow obstacle-free space, or obstacle configurations that are not in the training data set. 
Simulation results demonstrate the potential of diffusion models for reliable multi-agent formation planning.
\end{abstract}


\section{Introduction}

Multi-agent formation control is a fundamental problem in cooperative robotics, enabling robot teams to perform tasks such as environmental monitoring, search-and-rescue, and large-scale exploration. However, planning collision-free trajectories while maintaining formation remains a challenging task, especially in cluttered environments where the robots must avoid obstacles while preserving the formation.

Conventional planning approaches often rely on the ground-truth environment or the agents’ motion dynamics. Examples include trajectory optimization methods and model predictive control (MPC) \cite{Changyu2024tcst}, both of which rely on accurate dynamics models and environment structure. When successfully optimized, the resulting trajectories usually exhibit desired properties such as smoothness and optimized costs. However, they are sensitive to the initializations and may get stuck in local minima due to the non-convexity of the optimization problem. To handle these issues, learning-based techniques, like reinforcement learning \cite{KZhang2018icml}, train approximate dynamics models and incorporate them into planning procedures. However, these learned models are often exploited by planners, resulting in adversarial trajectories, poor generalization, and sensitivity to distribution shifts when moved in unseen environments. From these limitations, alternative methods are demanded for long-horizon, multimodal planning tasks in cluttered environments.

Recently, diffusion models have become a useful tool for generating long sequences of actions. First developed to model complex data distributions \cite{ho2020}, diffusion processes have demonstrated strong capabilities in trajectory generation, , handling multiple possible outcomes, and gradually refining solutions. \cite{janner2022} showed that diffusion-based planners can synthesize flexible long-horizon behaviors, while \cite{ajay2023} introduced differentiable trajectory optimization via diffusion for robotics applications. Based on these ideas, Diffusion Policy \cite{chi2024} generates full action sequences instead of predicting actions one step at a time, resulting in smoother and more consistent visuomotor control.

Motivated by these strengths, we use the Diffusion Policy framework \cite{chi2024} for multi-agent formation planning in cluttered environments.In our approach, a diffusion-based policy generates long action sequences for the formation reference point, defined as the midpoint between the leader agents. The policy takes recent state histories and a compact representation of obstacles as input, and gradually improves the action sequence through a denoising process.

Unlike autoregressive or single-step planners, this approach predicts an entire action sequence and progressively improves it via iterative denoising, resulting in smooth and temporally consistent trajectories while alleviating the local-minima issues commonly encountered by classical planning methods. This design leverages several key advantages of diffusion models:
\begin{itemize}
    \item \textbf{Multimodal action distributions:} Diffusion policies can represente multimodal action distributions instead of a single fixed solution.
    \item \textbf{Sequence prediction:} The diffusion model predicts entire action sequences rather than single-step actions, enabling long-horizon planning and avoiding short-sighted behavior.
    \item \textbf{Long-horizon planning:} Planning is a part of the iterative denoising process, allowing the model to refine actions over time and generate coherent trajectories.
    \item \textbf{Stable training:} Diffusion policies are trained using a denoising score matching objective, avoiding adversarial training and unstable optimization.
\end{itemize}

To achieve coordinated motion, we combine a distance-based formation tracking controller with a leader–follower structure. This design builds on earlier work in behavior-based formation control \cite{Balch}, bearing-based stabilization \cite{Trinh18}, and graph-theoretic multi-agent coordination \cite{Saber1}. The diffusion planner generates a reference trajectory for the formation leaders, which induces a time-varying desired formation assumed to be infinitesimally rigid in $\mathbb{R}^2$. Followers rely on measured relative positions to neighbors and apply a distributed control law consisting of two components: (i) a stabilization term that regulates inter-agent distance errors, and (ii) a tracking term that enables the formation to follow the leaders’ motion over time.

The remainder of this paper is organized as follows. Graph theory, the problem formulation and the Denoising Diffusion Probabilistic Model (DDPM) are given in Section \ref{sec:Method}. Section \ref{diffpath} introduces its application to path planning. The proposed formation tracking control law is given in Section~\ref{controllaw}. Simulation results are reported in Section~\ref{exp}, and concluding remarks are drawn in Section~\ref{conclu}.

\begin{figure*} 
\centering 
\includegraphics[width=1.0\textwidth]{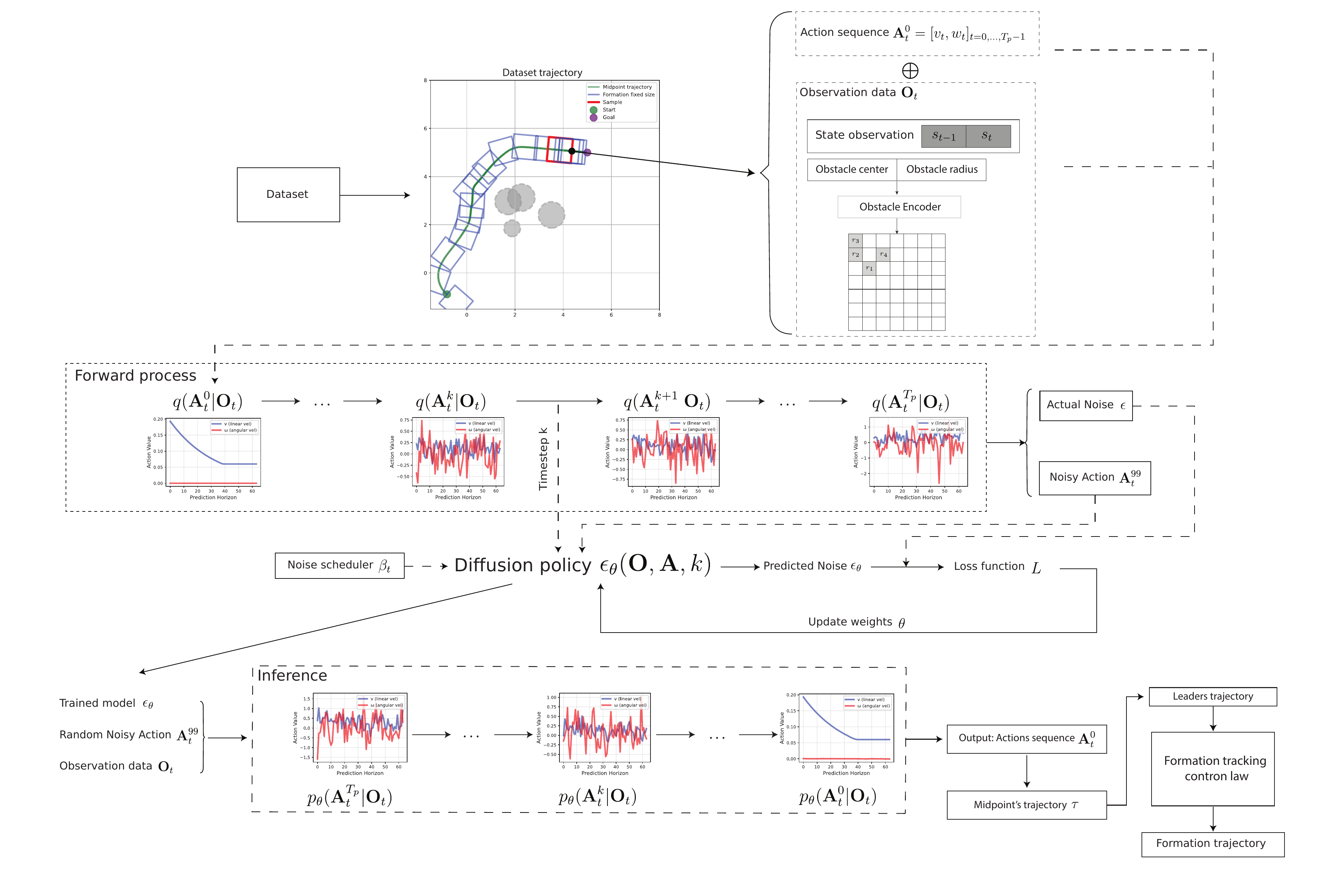} 
\caption{Overview of the proposed diffusion-based trajectory generation and formation tracking framework.} 
\label{overallfig} 
\end{figure*}

\section{Methodology}
\label{sec:Method}
\subsection{Graph Theory}
\label{graph}
The information flow of an $n$-agent system is described by a digraph $\mathcal{G} = (\mathcal{V},\mathcal{E})$ which has a set of vertices $\mathcal{V} = (1, 2,\ldots,n)$ and $\mathcal{E}\subseteq \mathcal{V} \times \mathcal{V} $. An edge in $\mathcal{E}$ is denoted by $e_{k} = (i,j), i, j \in \mathcal{V},k=1,2,\ldots,m$ with $m$ being the number of edges. $\mathcal{N}_i = \{ j | (i,j)\in \mathcal{E} \}$ is the neighborhood set of the agent $i$. Assign directions to the edges, the graph $\mathcal{G}$ has the incidence matrix $\mathbf{H} = [h_{ki}] \in \mathbb{R}^{m \times n}$ as $h_{ki} = 1 $ if $e_{k} = (j,i)$, $h_{ki} = -1 $ if $e_{k} = (i,j)$ and $h_{ki} = 0 $ otherwise. 

Let $\bm{p}_i \in \mathbb{R}^{dn}$ denote the positions of agent $i$ in global coordinate ${}^g\Sigma$. Additionally, $z_{ij} \triangleq \bm{p}_j - \bm{p} _i$ is denoted as the relative position in global frame. Along with the relative position, $d_{ij} \triangleq \lVert z_{ij} \rVert $ stands for the distance, $d_{ij}^*$ is the desired distance and $\epsilon_{ij} \triangleq d_{ij}^2 - d_{ij}^{*2}$ is the distance's error between 2 agents $i,j.$

Consider a formation $\left(\mathcal{G},\bm{p}\right)$ with the realization $\bm{p} \triangleq \operatorname{vec}(\bm p_1,\bm p_2,\dots,\bm p_n)$ and graph $\mathcal{G}$. The rigidity function of framework $\left(\mathcal{G},\bm{p}\right)$ is described as follows: 
\begin{equation}
\bm{f}_{\mathcal{G}}(\bm{p})\triangleq \left[ \ldots,\lVert z_{ij}\rVert^2,\ldots\right]^\intercal\in\mathbb{R}^m,\forall (i,j)\in\mathcal{E}
\label{r_G}
\end{equation}

Define $\mathbf{z} = \operatorname{vec}(z_1,\ldots,z_m)\in R^{2m}$, $\mathbf{z}=\bar{\mathbf{H}}\bm{p}$, where $\bar{\mathbf{H}}= \mathbf{H} \otimes \mathbf{I}_2$. The distance error vector $\bm{\varepsilon} = \operatorname{vec}(\bm{\varepsilon}_1,\ldots,\bm{\varepsilon}_m)$ and the rigidity matrix $\bm{R}$ are formulated as follows:

\begin{equation}
\begin{cases}
\bm{\varepsilon}_i \triangleq \lVert z_{ij}\rVert^2-(d_{ij}^*)^2,\forall j \in \mathcal{N}_i \\
\bm{R} \triangleq \frac{1}{2}\frac{\partial\bm{f}_{\mathcal{G}}}{\partial\bm{p}} = \mathbf{D}^{\intercal}(\mathbf{z)} \bar{\mathbf{H}} \in \mathbb{R}^{m\times 2n}
\label{eR}
\end{cases}
\end{equation}
where $\mathbf{D}(\mathbf{z)} = \operatorname{blkdiag}(z_i)_{i=1}^m \in \mathbb{R}^{2m \times m}$.  

\subsection{Problem Formulation}
\label{proform}
Consider an $n$-agent system containing $l \geq 2$ leaders and $f = n-l$ followers with a directed graph $\mathcal{G} = (\mathcal{V},\mathcal{E})$. The states of each agent are represented by the position vector $\bm{p}_i$ and the velocity $\bm{v}_i$ $\in \mathbb{R}^2$. The kinematic model of follower $i$ is described as follows \cite{TQuoc2020}

\begin{equation}
\begin{aligned}
& \dot{\bm{p}_i} = \bm{v}_i = \mathbf{h}_iu_i \\
& \dot{\mathbf{h}_i} = \omega_i \times \mathbf{h}_i  = -\mathbf{h}_i \times \omega_i
\end{aligned}
\label{dynafol}
\end{equation}
with $\mathbf{h}_i$ being the unit heading vector, $u_i$ the forward velocity along $\mathbf{h}_i$ and $\omega_i$ the angular velocity. 

Assume that the system has $n_l \geq 2$ leaders and the remaining $ n_f = n-n_l$ agents are called followers. The index sets of leader and follower agents are $\mathcal{V}_l = \{1, \ldots, n_l\}$ and $\mathcal{V}_f = \mathcal{V}\backslash \mathcal{V}_l$, respectively. The moving desired formation of the system is $\bm{p}^*(t)=\operatorname{vec}(\bm{p}_1^*(t),\bm{p}_2^*(t),\ldots,\bm{p}_n^*(t)) \in \mathbb{R}^{2n}$.

\begin{Assumption}\label{ass:leaders_paths} The leaders are initially positioned such that $\lVert \bm{p}_i - \bm{p}_j\rVert = \lVert \bm{p}_i^* - \bm{p}_j^*\rVert, \forall i,j \in \mathcal{V}_l $. They track the trajectory generated by Diffusion model. 
\end{Assumption}

\begin{Assumption}\label{ass:rigid_graph} The desired formation $(\mathcal{G}, \bm{p}^*)$ is infinitesimally bearing rigid.
\end{Assumption}

To evaluate the quality of the formation, we define the tracking error vectors $\bm{\delta}_p=\bm{p}(t)-\bm{p}^*(t)\in \mathbb{R}^{dn}$ and  $\bm{\delta}_{p_f}=\bm{p}(t)-\bm{p}^*(t)\in \mathbb{R}^{dn_f}$. Hence $\bm{\delta}_p=\operatorname{vec}(0_{dn},\bm{\delta}_{p_f})$ since $\bm{p}_i(t)=\bm{p}^*_i(t),\forall i \in \mathcal{V}_l$.

\textit{Problem:} Under these \textit{Assumptions}, design a distributed control law $(\bm{u}_i,\bm{\omega}_i),\forall i \in \mathcal{V}_f$ so that $\bm{\delta}_p \rightarrow0$.

\subsection{Denoising Diffusion Probabilistic Models (DDPM)}
\label{ddpm}

According to \cite{ho2020}, \cite{sohl2015}, denoising diffusion probabilistic models are generative models that transform a simple known distribution, such as a Gaussian distribution, into a target distribution through a parameterized Markov diffusion process over a finite number of steps, enabling exact sample generation and extreme structural flexibility. Generally, DDPMs add Gaussian noise to target sample ${\mathbf{x}}^0$ through diffusion process (forward process) and train a neural network to reverse the diffusion process for sample generation.

In \cite{ho2020}, forward process or diffusion process is a Markov chain that gradually adds Gaussian noise to given targeted sample ${\mathbf{A}}^0$ according to a variance schedule $\beta_0, \beta_1,\ldots,\beta_T$: 
\begin{equation}
q(\mathrm{\mathbf{A}}^{1:T}|\mathrm{\mathbf{A}}^0) \; \triangleq \; \prod_{t=1}^T q(\mathrm{\mathbf{A}}^k|\mathrm{\mathbf{A}}^{k-1})
\end{equation}
\begin{equation}
q(\mathrm{\mathbf{A}}^k|\mathrm{\mathbf{A}}^{k-1}) \; \triangleq \; \mathcal{N}\left(\mathrm{\mathbf{A}}^k;\sqrt{1-\beta_k}\,\mathrm{\mathbf{A}}^{k-1}, \beta_k \mathbf{I}\right)
\label{beta}
\end{equation}

However, \eqref{beta} requires $k-1$ sequential transitions to sample $\mathbf{A}^k$, leading to inefficiency. To overcome this limitation, $\mathbf{A}^k$ can be sampled directly from $\mathbf{A}^0$:

\begin{equation}
q(\mathrm{\mathbf{A}}^k|\mathrm{\mathbf{A}}^{0}) \; \triangleq \; \mathcal{N}\left(\mathrm{\mathbf{A}}^k;\sqrt{\overline{\alpha}_k}\,\mathrm{\mathbf{A}}^{0}, (1-\overline{\alpha}_k)\mathbf{I}\right)
\label{qalpha}
\end{equation}
where $\alpha_k \triangleq 1 - \beta_k$ and $\overline{\alpha}_k \triangleq \prod_{s=1}^k \alpha_s$.

Furthermore, we can perform the forward process under the linear expression:
\begin{equation}
\mathrm{\mathbf{A}}^k(\mathrm{\mathbf{A}}^0,\boldsymbol{\epsilon}) = \sqrt{\overline{\alpha}_k}\mathrm{\mathbf{x}}^0 +  \sqrt{1-\overline{\alpha}_k}\boldsymbol{\epsilon}
\label{forlin}
\end{equation}
where $\epsilon \in \mathcal{N}(0,\mathbf{I})$ is the random noise sampled from Gaussian noise. 

The forward process converges to a standard Gaussian distribution. The reverse process samples from this distribution, with the neural network predicting the parameters.

\begin{equation}
p_\theta(\mathrm{\mathbf{A}}^{0:T}) \; \triangleq \; p(\mathrm{\mathbf{A}}^T) \prod_{k=1}^T p_\theta(\mathrm{\mathbf{A}}^{k-1}|\mathrm{\mathbf{A}}^{k})
\end{equation}
\begin{equation}
p_\theta(\mathrm{\mathbf{A}}^{k-1}|\mathrm{\mathbf{A}}^{k}) \; \triangleq \; \mathcal{N}\left(\mathrm{\mathbf{A}}^{k-1};\bm{\mu}_\theta(\mathrm{\mathbf{A}}^k,k), \beta_k \mathbf{I}\right)
\label{rev}
\end{equation}
where $\bm{\mu}_\theta$ is mean of the Gaussian distribution, which is predicted by a neural network to model the distribution of $\mathbf{A}^{k-1}$.

Sampling $\mathrm{\mathbf{A}}^{k-1} \sim p_\theta(\mathrm{\mathbf{A}}^{k-1}|\mathrm{\mathbf{A}}^{k}) $ is equivalent to computing: 
\begin{equation}
\mathbf{A}^{k-1} = \frac{1}{\sqrt{\alpha_k}} \left( \mathbf{A}^k - \frac{\beta_k}{\sqrt{1 - \bar{\alpha}_k}} \, \boldsymbol{\epsilon}_\theta(\mathbf{A}^k, k)\right)
+ \sigma_k \mathbf{z}
\label{simsam}
\end{equation}

Training aims to optimize $\theta$ by minimizing a variational bound $L$ on the negative log-likelihood: $\theta^* = \arg\min_\theta -\mathbb{E}[\log p_\theta(\mathrm{\mathbf{A}}^0)] $. The objective can be expressed by reformulating this bound and discarding constant terms:

\begin{equation}
  L \triangleq  \mathbb{E}_q \left[ \frac{1}{2\sigma_k^2} \lVert \Tilde{\bm{\mu}}_k(\mathrm{\mathbf{A}}^k,\mathrm{\mathbf{A}}^0) - \bm{\mu}_\theta(\mathrm{\mathbf{A}}^k,\mathrm{\mathbf{A}}^0) \rVert \right]
\label{lossdp}
\end{equation}

By replacing \eqref{lossdp} with \eqref{forlin} and applying algebraic transformations, \cite{ho2020} showed that the loss function can be simplified as :

\begin{equation}
L = \mathbb{E}_{k,\mathbf{A}_0,\boldsymbol{\epsilon}}\left[ \lVert \boldsymbol{\epsilon} - \boldsymbol{\epsilon}_\theta \rVert ^2 \right]
\label{simpleloss}
\end{equation}

\section{Diffusion for path planning}
\label{diffpath}	
\subsubsection{Path planning}

In this work, our approach is mainly inspired by the diffusion-based framework introduced in \cite{chi2024} and \cite{lu2024}, where diffusion model is adapted to sequential decision-making tasks.  

Our approach predicts the trajectory of the midpoint between the two leaders rather than individual leader trajectories. This design choice will reduce the learning complexity and enforces that both leaders move in a coordinated manner as a rigid bar, naturally avoiding scenarios where one leader navigates successfully with the obstacles.

Let $s^m = \left[\bm{q}^m,\dot{\bm{q}}^m\right]^\intercal \in \mathbb{R}^d $ be the state of the midpoint of the line connecting the two leaders, where $\bm{q}^m = \left[\bm{p}^m,\phi^m\right]^\intercal$ is the position vector and $\dot{\bm{q}}^m = [ \dot{\bm{p}}^m,\omega^m ]^\intercal$  is the velocity vector.

The discrete-time dynamics are:  
  
\begin{equation}
    \bm{q}^m_{t+1}= \bm{q}^m_{t} +  \begin{bmatrix} v_t\cos(\phi_t) \\ v_t\sin(\phi_t) \\ \omega_t\end{bmatrix} \Delta t  = \bm{q}^m_{t} + \begin{bmatrix}
        \cos(\phi_t) & 0 \\ \sin(\phi_t) &0 \\ 0 &1
    \end{bmatrix}a_t\Delta_t
\end{equation}   
where $a_t = [v_t, \omega_t]^\top$ is the velocity command applied at the midpoint.

The midpoint's trajectory is represented as an action sequence $\mathbf{A}_t \triangleq (a_t,a_{t+1},\ldots,a_{t+T_{p}-1}) \in \mathbb{R}^{T_P\times d_a} $ over a predicting horizon $T_p$. Following \cite{chi2024}, to achieve temporal consistency and smoothness in trajectory prediction, the diffusion model combined with Receding Horizon Control \cite{70083} is used to predict actions sequence on the observation data $\mathbf{O}_t$, while only the first $T_a$ steps of actions are executed.

The observation $\mathbf{O}_t$ contains the latest $T_0$ midpoint states $s^m$, the obstacle encoding $Obs$. To be more specific, the obstacle encoding is presented as a flattened vector from a matrix $Obs \in \mathbb{R}^{7\times7}$. For each grid cell
$(i,j)$ where $i,j \in \{1,2,\ldots,6\}$, the obstacle center $c_k$ is mapped to cell $(i,j)$ via $(i,j) = \lfloor c_k \rfloor - (1,1)$ and its information contains the radius of that obstacle $Obs[i,j]=r_k$.

In \cite{chi2024}, actions are predicted from the conditional distribution $p(\mathbf{A}_t|\mathbf{O}_t)$, avoiding explicit future states inference and improving both efficiency and accuracy. 

To adapt the current model, \cite{chi2024} has modified \eqref{simsam} and \eqref{simpleloss} as
\begin{equation}
    \mathbf{A}_t^{k-1} = \alpha\left(\mathbf{A}_t^k - \gamma \epsilon_\theta\left(\mathbf{O}_t,\mathbf{A}_t^k,k \right) + \mathcal{N}\left(0,\sigma^2\mathbf{I}\right)\right)
    \label{sampling}
\end{equation}
\begin{equation}
    L = MSE\left(\epsilon^k,\epsilon_\theta\left(\mathbf{O}_t,\mathbf{A}_t^k,k \right)\right)
    \label{loss}
\end{equation}

where $\alpha,\gamma,\sigma$ called the noise scheduler and the additive Gaussian Noise $\epsilon^k$ presented as a function of timestep $k$, have already been studied in \cite{ho2020}, \cite{nichol2021}. 

\subsubsection{Neural Network Architecture}
In this work, we adopt the CNN-based diffusion policy from \cite{chi2024} to implement the model $\epsilon_\theta$.

The network inputs consists of a noisy action trajectory $\mathbf{A}_t^{k}$, the diffusion step embedding and a global conditioning vector $\mathbf{O}_t$ containing normalized state and goal information. 

The diffusion step $k$ is encoded by using Sinusoidal positional embeddings \cite{attentionneed}, which provide a continuous vector representation of the timestep $k$ and allow the model to capture the relative positions between diffusion steps, thereby improving temporal awareness during the denoising process.

Furthermore, we adopt the Feature-wise Linear Modulation (FiLM) \cite{film}, following \cite{chi2024}, to adapt the conditioning vector $\mathbf{O}_t$ into the network. In each Conditional Residual Block, $\mathbf{O}_t$ is encoded to produce feature-wise scale $\gamma$ and bias $\beta$ components, which are then applied to the feature activations using FiLM modulation:

\begin{equation}
    \mathbf{y} = \gamma(\mathbf{O}_t) \odot \mathbf{x} + \beta(\mathbf{O}_t)
\end{equation}
where $\mathbf{x}$ and $\mathbf{y}$ denote the input and modulated feature maps, respectively. This modulation mechanism enables the network to adapt its feature activations to different task-specific conditions and environmental contexts.

The backbone of CNN-based diffusion policy in \cite{chi2024} was designed based on the U-NET architecture in \cite{janner2022}, \cite{cnnunet}, with some modifications to better accommodate temporal action data. Specifically, 1D convolutional layers are used instead of 2D convolutions to process sequential action features over time.

We adopt this architecture due to its ability to learn hierarchical representations from noisy action sequences and the environmental observations, where higher-level features are built upon lower-level ones, using an encoder–decoder structure with skip connections, which effectively adapts the absolute and relative aspects of action sequences. Furthermore, CNNs also offer stable training and can model multimodal behaviors in complex environments.

However, as noted in \cite{chi2024}, CNN-based U-NET architecture struggles with rapidly changing action sequences, due to the inductive bias of temporal convolutions favoring low-frequency signals. Another limitation of CNN is its restricted receptive field which makes difficult to capture the long-range dependencies. Finally, although CNN is easy to train, more complex architectures require a large amount of training data to generalize effectively.

\subsubsection{Training and Sampling process}

The procedures of training and inference are summarized in Algorithm~\ref{algtrain} and Algorithm~\ref{algsampling}, respectively.

\begin{algorithm}
\SetAlgoLined
\KwIn{Processed dataset $\mathcal{D} = \{(\mathbf{O}_t, \mathbf{A}_t)\}$, number of epochs $E$, batch size $B$, learning rate $\eta$, network $\epsilon_\theta$} 
\KwOut{Trained parameters $\theta^*$}
\For{epoch $1$ \KwTo $E$}{
\For{each batch $\{(\mathbf{O}^{(i)}, \mathbf{A}^{(i)})\}_{i=1}^B$ in $\mathcal{D}$}{
Sample noise $\boldsymbol{\epsilon} \sim \mathcal{N}(0, \mathbf{I}) \in \mathbb{R}^{B \times T_p \times d_a}$\;  
Sample random timestep $k \sim \text{Uniform}\{0, 1, \ldots, K-1\}$\;
Add random noise $\mathbf{A}^k = \sqrt{\bar{\alpha}_k} \mathbf{A} + \sqrt{1-\bar{\alpha}_k} \boldsymbol{\epsilon}$\;  
Predict noise $\hat{\boldsymbol{\epsilon}} = \epsilon_\theta(\mathbf{A}^k, k, \mathbf{O})$\;
Compute loss function $L = \text{MSE}(\hat{\boldsymbol{\epsilon}}, \boldsymbol{\epsilon})$\; 
Gradient update $\theta \leftarrow \theta - \eta \nabla_\theta \mathcal{L}$\;  
}
}
\caption{Training algorithm}
\label{algtrain}
\end{algorithm}

\begin{algorithm}
\SetAlgoLined
\KwIn{Start states $s^m_t$, obstacles encoding $Obs$, trained model $\epsilon_\theta$, noise scheduler $\beta$}
\KwOut{Action sequence $A^{k-1}_t$}
\While{not done}{
Normalize latest $T_o$ states $\tilde{s}^m_t$\;
Observation data: $\mathbf{O}_t = [\text{flatten}(\tilde{s}^m_t), Obs]$\;  
Noisy action sequence: $\mathbf{A}^K_t \sim \mathcal{N}(0, \mathbf{I}) \in \mathbb{R}^{T_p \times d_a}$\;    \For{$k = K-1$ \KwTo $0$}{  
        Predict noise: $\hat{\boldsymbol{\epsilon}} = \epsilon_\theta(\mathbf{A}^k_t, k, \mathbf{O})$\;            
        Denoise: $\mathbf{A}^{k-1}_t = \text{scheduler.step}(\hat{\boldsymbol{\epsilon}}, k, \mathbf{A}^k_t)$\;  
    }    
Unnormalize $\mathbf{A}^0_t$\;
Extract action: $\mathbf{A}^e_t = \mathbf{A}^0[T_0-1:T_0-1+T_a, :]$\;
\For{$i = 0$ \KwTo $T_a - 1$}{  
    Execute action $\mathbf{A}^e_i$ on system\;  
    Update states $s^m_{t+i+1}$\;  
    Add to observation history\;  
}
}
\caption{Sampling algorithm}
\label{algsampling}
\end{algorithm}

\section{DISTANCE-BASED FORMATION TRACKING CONTROL}
\label{controllaw}

Define $\bm{\chi}_i = \sum_{j\in\mathcal{N}_i}{(d_{ij}^2 - d^{*2}_{ij})(\bm{p}_i-\bm{p}_j)} \in \mathbb{R}^2$.
The control law for each follower $i \in \mathcal{V}_f$ is proposed as follows:

\begin{equation}
    \begin{cases}
        \bm{u}_i = \mathbf{h}_i^{\intercal}(k_1\bm{\chi}_i+\beta \operatorname{sign}(\bm{\chi}_i)) \\
        \omega_i = \mathbf{h}_i \times (k_2\bm{\chi}_i+\beta \operatorname{sign}(\bm{\chi}_i))
    \end{cases}
    \label{fol1}
\end{equation}
where $k_1,k_2,\beta > 0$ are constant control gains.

Under the control law \eqref{fol1}, the closed-loop system of follower $i$ is obtained as:

\begin{equation}
    \begin{cases}
        \dot{\bm{p}}_i = \mathbf{h}_i\mathbf{h}^\intercal_i(k_1\bm{\chi}_i+\beta \operatorname{sign}(\bm{\chi}_i))\\
        \begin{aligned}
            \dot{\mathbf{h}}_i &= \mathbf{h}_i\times\mathbf{h}_i\times(k_2\bm{\chi}_i+\beta \operatorname{sign}(\bm{\chi}_i))\\
            &=  (\mathbf{I}_d-\mathbf{h}_i\mathbf{h}_i^{\intercal})(k_2\bm{\chi}_i+\beta \operatorname{sign}(\bm{\chi}_i))
            
        \end{aligned}
        
    \end{cases}
\end{equation}

In \cite{bearingtrack}, the following stacked vectors and matrices are defined:

\begin{equation}
    \begin{aligned}
        & \bm{Z} =  \operatorname{blkdiag}(0_{dn_l\times dn_l},\mathbf{I}_{dn_f \times dn_f}) \in \mathbb{R}^{dn \times dn},\\
        & \mathbf{h} = \operatorname{vec}(\mathbf{h}_1,\ldots,\mathbf{h}_n) \in \mathbb{R}^{dn}, \\
        & \bm{D}_{\mathbf{h}_i} = \operatorname{blkdiag}(\{\mathbf{h}_i\mathbf{h}_i^{\intercal}\}_{i\in \mathcal{V}}) \in \mathbb{R}^{dn \times dn},\\
        & \bm{D}_{\mathbf{h}_i^\intercal} = \operatorname{blkdiag}(\{\mathbf{I}_d -\mathbf{h}_i\mathbf{h}_i^{\intercal}\}_{i\in \mathcal{V}}) \in \mathbb{R}^{dn \times dn}.\\
        & \bm{\chi} = \operatorname{vec}( \bm{\chi}_1, \ldots, \bm{\chi}_m) = \mathbf{R}^\intercal\bm{\varepsilon} \in \mathbb{R}^{2m}
    \end{aligned}
\end{equation}

Using these definitions, the closed-loop dynamics of the $n$-agent system can be written as

\begin{equation}
    \begin{aligned}
        \dot{\bm{p}} &= \operatorname{col}(v_1,\ldots,v_{n_l},0_{dn_f}) - \\&\operatorname{blkdiag}(0_{dn_l\times dn_l},\{\mathbf{h}_i\mathbf{h}_i^{\intercal}\}_{i\in \mathcal{V}}) (k_1\bm{\chi}+\beta \operatorname{sign}(\bm{\chi})) \\
        &= \operatorname{col}(v_1,\ldots,v_{n_l},0_{dn_f}) - Z\bm{D}_{\mathbf{h}_i} (k_1\bm{\chi}+\beta \operatorname{sign}(\bm{\chi}))\\
        \dot{\mathbf{h}} &= \operatorname{col}(\omega_1,\ldots,\omega_{n_l},0_{dn_f}) \times \mathbf{h}-\\
        &\operatorname{blkdiag}(0_{dn_l\times dn_l},\{\mathbf{I}_d -\mathbf{h}_i\mathbf{h}_i^{\intercal}\}_{i\in \mathcal{V}}) \times(k_2\bm{\chi}+\beta \operatorname{sign}(\bm{\chi})) \\
        &=\operatorname{col}(\omega_1,\ldots,\omega_{n_l},0_{dn_f}) \times \mathbf{h} -Z\bm{D}_{\mathbf{h}_i^\intercal}(k_2\bm{\chi}+\beta \operatorname{sign}(\bm{\chi}))
    \end{aligned}
\end{equation}

where $v_1,\ldots,v_n$ and $\omega_1,\ldots,\omega_n$ are the forward velocities and angular velocities generated by diffusion path planning for the leaders, respectively.

\section{EXPERIMENTS}
\label{exp}
\subsection{Experimental setup}

Consider a $4$-agent system that consists of $2$ leaders and $2$ followers, with the desired distance vector is defined as $\bm {f}_{\mathcal{G}}^* = [1,2,1,1,1]$ .The leaders' trajectories are generated based on the midpoint's trajectory predicted by the diffusion model. The dataset contains $1878$ trajectories collected in environments containing three to five circular obstacles randomly placed within a $[1.5,1.5] \times [6,6]$ workspace, with obstacle radii ranging from $0.3$ to $0.8$ meters.

The initial positions of the agents are selected as $\bm{p}_{1l}(0)=[-0.5,0],\bm{p}_{2l}(0)=[0.5,0],\bm{p}_{1f}(0)=[-0.2,-0.9],\bm{p}_{2f}(0)=[0.4,-1]$. The initial headings are  $\phi_{1l}(0)=\phi_{2l}(0)=\phi_{2f}(0)=90^{\degree},\phi_{1f}(0)=180^{\degree} $. 

Each episode runs for at most $T_{\max}=600$ steps with a sampling step being $\Delta t = 0.1\,$s. 

To generate training trajectories, we use a hand-crafted Path-Aware Controller (PAC), which is a reactive navigation controller based on artificial potential fields. The controller combines a goal-oriented attractive component with  repulsive forces for obstacle avoidance, and checks whether the direct heading path to the goal is blocked. When obstacles block the heading path, additional lateral repulsive forces are used to encourage safe detours rather than backward motion. The resulting force vector is converted into linear and angular velocity commands. Importantly, this controller is used solely to collect demonstration trajectories for training the diffusion policy and is not part of the deployed system.

The controller gains for the tracking law are set to $k_1 = 35$, $k_2 = 30$, and $\beta = 23$.

The training hyperparameters are summarized in Table~\ref{tab:training_hyperparams}.

\begin{table*}[t]
\centering
\caption{Training and Model Hyperparameters}
\label{tab:training_hyperparams}
{
\begin{tabular}{l| r}
\toprule
\textbf{Category} & \textbf{Description}  \\
\midrule

\multicolumn{2}{l}{\textbf{Training Hyperparameters}} \\
\hline
Batch size & 256\\
Optimizer & AdamW with learning rate $1.05\times10^{-4}$ and weight decay $1\times10^{-6}$ \\
Learning rate scheduler & Cosine annealing with 1000 warmup steps \\
EMA (Exponential Moving Average) & Enabled with power = 0.75 for model weight averaging \\

\midrule
\multicolumn{2}{l}{\textbf{Diffusion Model Hyperparameters}} \\
\hline
Diffusion timesteps & 100  \\
Beta schedule & \texttt{squaredcos\_cap\_v2} (cosine-based variance schedule)  \\
Prediction type & \texttt{epsilon} (model predicts noise $\varepsilon_\theta$) \\
Clip sample & True (for numerical stability) \\

\midrule
\multicolumn{2}{l}{\textbf{Network Architecture Hyperparameters}} \\
\hline
Step embedding dimension & 256  \\
Down dimensions & [256, 512, 1024]  \\
Kernel size & 5 \\
GroupNorm groups & 8 \\

\midrule
\multicolumn{2}{l}{\textbf{Temporal Hyperparameters}} \\
\hline
Prediction horizon ($T_p$) & 64 timesteps \\
Observation horizon ($T_o$) & 2 timesteps\\
Action horizon ($T_a$) & 10 timesteps  \\

\bottomrule
\end{tabular}
}
\end{table*}

\subsection{Experiment results}

We first evaluate the proposed diffusion-based planner against two representative baselines:  a Model Predictive Path Integral (MPPI) \cite{mppi} controller, and path-aware controller (PAC). All methods are evaluated on the same set of $100$ dataset episodes to ensure a fair comparison.

The performance of each method is evaluated through multiple navigation and control metrics, including \textbf{success rate}, \textbf{navigation efficiency}, \textbf{collision safety}, \textbf{control cost}, \textbf{speed performance}, and \textbf{motion quality}. Arrows show whether a higher ($\uparrow$) or lower ($\downarrow$) value corresponds to better performance. The quantitative results are summarized in Table~\ref{tab:controller_comparison_dataset}.

\begin{table*}[htbp]
\centering
\caption{Three-Way Controller Comparison on Dataset: MPPI vs Diffusion vs PAC (100 Episodes)}
\label{tab:controller_comparison_dataset}
{
\begin{tabular}{llccc}
\toprule
\textbf{Category} & \textbf{Metric} & \textbf{MPPI} & \textbf{Diffusion} & \textbf{PAC} \\
\midrule

\multirow{1}{*}{\textbf{SUCCESS}}
& Success Rate (\%) $\uparrow$ & $65.0$ & $62.0$ & $\bm{100.0}$ \\

\midrule
\multirow{3}{*}{\textbf{NAVIGATION EFFICIENCY}}
& Path Length (m) $\downarrow$ & $9.35 \pm 1.52$ & $\bm{9.12 \pm 1.33}$ & $9.38 \pm 1.83$ \\
& Path Optimality $\uparrow$ & $0.775 \pm 0.120$ & $\bm{0.791 \pm 0.110}$ & $0.780 \pm 0.133$ \\
& Tracking Deviation (m) $\downarrow$ & $1.067 \pm 0.447$ & $\bm{1.064 \pm 0.419}$ & $1.139 \pm 0.513$ \\

\midrule
\multirow{3}{*}{\textbf{COLLISION SAFETY}}
& Min Clearance (m) $\uparrow$ & $0.440 \pm 0.193$ & $0.305 \pm 0.237$ & $\bm{0.488 \pm 0.140}$ \\
& Mean Clearance (m) $\uparrow$ & $0.974 \pm 0.164$ & $0.895 \pm 0.174$ & $\bm{1.078 \pm 0.180}$ \\

\midrule
\multirow{3}{*}{\textbf{CONTROL COST}}
& Mean Control Magnitude $\downarrow$ & $0.602 \pm 0.079$ & $\bm{0.370 \pm 0.031}$ & $0.452 \pm 0.050$ \\
& Control Smoothness $\downarrow$ & $4.538 \pm 1.151$ & $\bm{0.336 \pm 0.102}$ & $1.164 \pm 0.650$ \\
& Energy Usage $\downarrow$ & $10.20 \pm 3.11$ & $\bm{5.31 \pm 1.34}$ & $6.95 \pm 2.00$ \\

\midrule
\multirow{2}{*}{\textbf{SPEED PERFORMANCE}}
& Time (s) $\downarrow$ & $\bm{25.98 \pm 10.20}$ & $31.05 \pm 6.75$ & $27.81 \pm 6.49$ \\
& Mean Velocity (m/s) $\uparrow$ & $\bm{0.390 \pm 0.084}$ & $0.297 \pm 0.023$ & $0.340 \pm 0.040$ \\

\midrule
\multirow{4}{*}{\textbf{MOTION QUALITY}}
& Mean Curvature $\downarrow$ & $28.29 \pm 52.73$ & $0.787 \pm 0.606$ & $\bm{0.733 \pm 0.388}$ \\
& Peak Curvature $\downarrow$ & $6767 \pm 15229$ & $\bm{5.28 \pm 2.77}$ & $5.51 \pm 3.57$ \\
& Motion Jerk $\downarrow$ & $23.10 \pm 3.51$ & $\bm{0.359 \pm 0.055}$ & $0.576 \pm 0.114$ \\
& Orientation Stability $\downarrow$ & $0.202 \pm 0.028$ & $\bm{0.188 \pm 0.023}$ & $0.225 \pm 0.026$ \\

\bottomrule
\end{tabular}
}
\end{table*}

As shown in Table~\ref{tab:controller_comparison_dataset}, PAC achieves a $100\%$ success rate since the dataset was generated using a similar reactive strategy. In contrast, the diffusion-based planner and MPPI achieve success rates of $62\%$ and $65\%$, respectively.

Even though it doesn’t always succeed, the diffusion
planner makes paths that are smoother and use less energy. It has the lowest control, smoothness, and energy use. It is even better than MPPI and PAC in terms of mean curve, peak curve, and jerky movement. Further, it finds the shortest paths more efficiently, with all paths staying closer to the intended route than that of MPPI and PAC.

Although the PAC controller demonstrated the finest obstacle clearance, it increased safety margin and comes at the expense of higher control effort and reduced motion smoothness compared to the diffusion-based planner. Overall, the diffusion-based approach is capable of producing efficient trajectories.

\begin{table}[ht]
\centering
\caption{Standard vs Adaptive Diffusion Policy}
\label{tab:succ_fail}
\begin{tabular}{llrr}
\hline
\textbf{Dataset} & \textbf{Method} & \textbf{Success (\%)} & \textbf{Failure (\%)} \\ \hline
Random seeds & Standard & 25\% & 75\% \\ 
Random seeds & Adaptive & 26\% & 74\% \\ 
Dataset episodes & Adaptive & 70\% & 30\% \\ \hline
\end{tabular}
\end{table}

We evaluate the performance of the diffusion policy in unseen environments by comparing a \emph{standard} single-sample setting with an \emph{adaptive} variant that samples 10 candidate action sequences and selects the safest path. The success and failure rates are reported in Table~\ref{tab:succ_fail}.

In validation environments, the adaptive variant slightly improves the success rate from $25\%$ to $26\%$ compared to the standard single-sample setting. In contrast, when evaluated on dataset episodes, the adaptive diffusion policy achieves a higher success rate of $70\%$ due to its strong performance in familiar environments.

To criticize the consensus performance of the control law, we analyze two representative successful seeds and analyze the distribution of formation error and the velocities of the followers' tracking. 

\begin{figure*}[ht]
    \centering
    \begin{subfigure}[b]{0.9\textwidth}
        \centering
        \includegraphics[width=\textwidth]{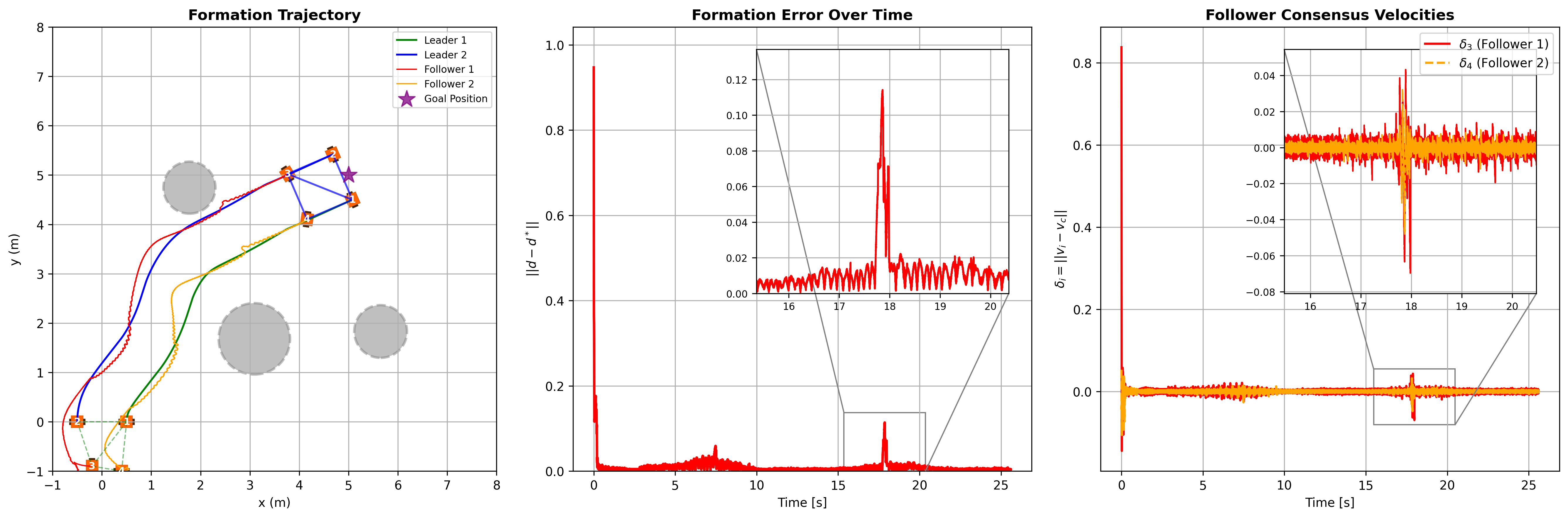}
        \caption{Trajectory 1}
        \label{traj1}
    \end{subfigure}
    \vspace{1em} 

    \begin{subfigure}[b]{0.9\textwidth}
        \centering
        \includegraphics[width=\textwidth]{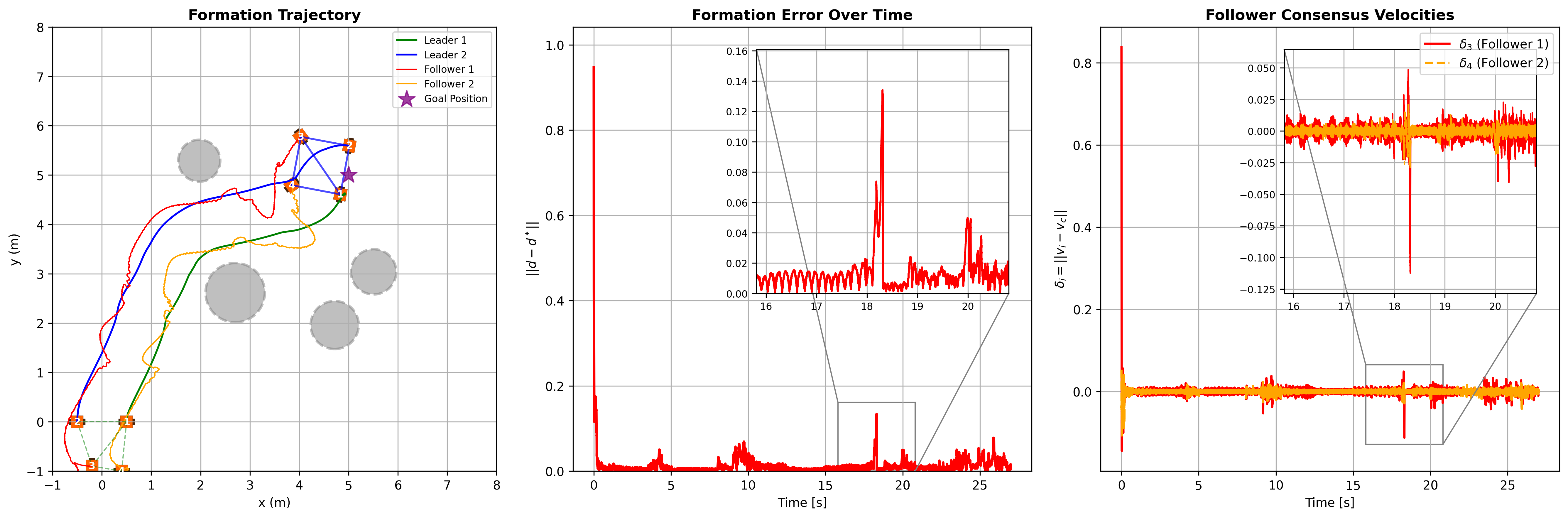}
        \caption{Trajectory 2}
        \label{traj2}
    \end{subfigure}
    \vspace{1em}
    
    \begin{subfigure}[b]{0.9\textwidth}
        \centering
        \includegraphics[width=\textwidth]{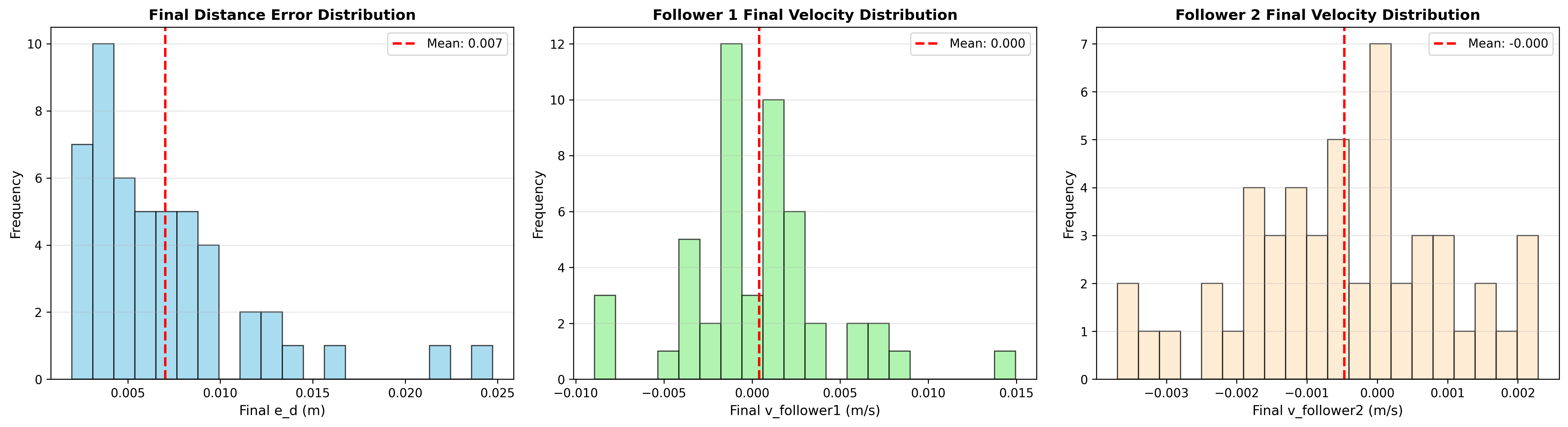}
        \caption{Final Tracking Error Distribution}
        \label{histogram}
    \end{subfigure}
    \vspace{1em}

    \caption{Formation trajectories in different environments.}
    \label{3fig}
\end{figure*}

In Fig.~\ref{3fig}, the predicted leaders trajectories are smooth and the followers can maintain the formation while avoiding obstacles. The formation error $e_d = \lVert d-d^* \rVert$ converges to $0$ over time but oscillations still occur when avoiding obstacles. Similar oscillations are observed in the followers’ velocity errors. Statistics collected over $50$ successful trajectories presented in Fig.~\ref{histogram} show that both formation error and the followers' velocity errors approach near zero when time increases.

\begin{figure*} 
\centering 
\includegraphics[width=.7\textwidth]{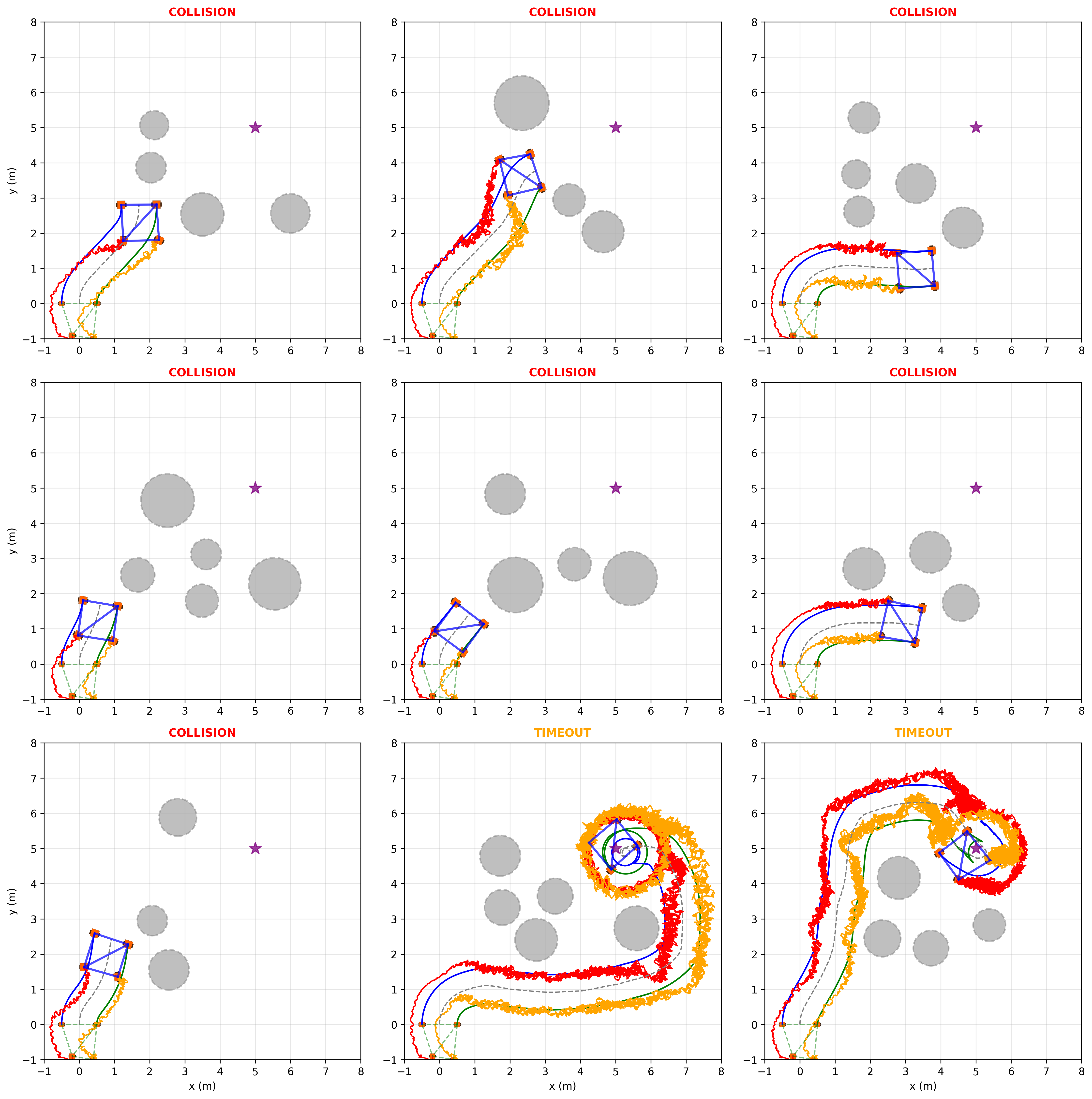 } 
\caption{Representative failure cases} 
\label{9fig} 
\end{figure*}

Fig.~\ref{9fig} shows 9 failure cases (7 collisions and 2 timeouts in this sample). Most failures are caused by collisions in narrow or cluttered environments. The timeout cases occur when the midpoint stuck near the goal, leading to oscillatory actions. 

From these observations, we identify two main causes for the low success rate. First, the "out-of-distribution" might be one of the main reasons, as the training dataset was generated by a reactive controller that itself achieves only around $50\%$ success and does not cover sufficiently complex environments. Second, the CNN-based diffusion policy oversmooths action sequences, leading to a lack of the curvature required for sharp turns in cluttered environments. As a result, the planner produces smooth motions with good quality metrics but struggles with handling cluttered scenes that require aggressive avoidance actions. The limited receptive field of CNNs also prevents the model from considering long-range spatial constraints, such as narrow passages.

\section{CONCLUSION AND FUTURE WORK}\label{conclu}

In this paper, we proposed a diffusion-based planning pipeline for leader–follower formation control, where the diffusion policy predicts short action sequences for the midpoint, which are then transformed to leaders' trajectories. The followers maintain the formation by using a distributed distance-based tracking controller. The approach generates smooth leader motions and allows followers to reduce distance and velocity errors to nearly zero.

Analysis of successful runs shows fast convergence of formation and velocity errors, while failures mainly result from collisions in narrow or cluttered environments.

To overcome the limitations, we plan to pursue the following directions for future work:

\begin{itemize}
  \item \textbf{Data and robustness:} expand and vary the training dataset with harder environment, and apply augmentation or curriculum learning to improve out-of-distributions generalization.
  \item \textbf{Model and planning improvements:} use Transformer- or attention-based diffusion backbones and increase the prediction horizon $T_P$ to plan farther ahead.
  \item \textbf{Safety and verification:} add a safety layer that rejects or adjust unsafe action before execution.
  \item \textbf{Controller refinement and deployment:} refine more adaptive or robust formation controllers and test the full pipeline on real robots or higher-fidelity simulators.
\end{itemize}

With these proposed improvements, collision and timeout cases are expected to decrease, thereby achieving higher robustness. Overall, this work demonstrates the promise of diffusion-based planning for multi-agent formation tasks.

\bibliographystyle{IEEEtran}
\bibliography{References}

\end{document}